\documentclass{article}
\usepackage{spconf,amsmath,graphicx}
\usepackage{booktabs}
\usepackage{amsmath}
\usepackage{amssymb}
\usepackage{multirow}
\usepackage[pagebackref=true,breaklinks=true,letterpaper=true,colorlinks,bookmarks=false]{hyperref}

\title{BANDWIDTH-EFFICIENT INFERENCE FOR NERUAL IMAGE COMPRESSION}
\name{Shanzhi Yin\textsuperscript{12$\ast$},
\thanks{$\ast$This work was done when Shanzhi Yin was intern at Institute for AI Industry Research (AIR), Tsinghua University.}
Tongda Xu\textsuperscript{1}, Yongsheng Liang\textsuperscript{2},Yuanyuan Wang\textsuperscript{3}, Yanghao Li\textsuperscript{1}, Yan Wang\textsuperscript{1$\dagger$},
\thanks{$\dagger$Yan Wang is the corresponding author.}
Jingjing Liu\textsuperscript{1}}
\address{Institute for AI Industry Research (AIR), Tsinghua University\textsuperscript{1}\\
 Harbin Institute of Technology (Shenzhen)\textsuperscript{2} \\
SenseTime Research\textsuperscript{3}\\}
\begin{document}
%
\maketitle
\begin{abstract}
With neural networks growing deeper and feature maps growing larger, limited communication bandwidth with external memory (or DRAM) and power constraints become a bottleneck in implementing network inference on mobile and edge devices. In this paper, we propose an end-to-end differentiable bandwidth efficient neural inference method with the activation compressed by neural data compression method. Specifically, we propose a transform-quantization-entropy coding pipeline for activation compression with symmetric exponential Golomb coding and a data-dependent Gaussian entropy model for arithmetic coding. Optimized with existing model quantization methods, low-level task of image compression can achieve up to 19$\times$ bandwidth reduction with 6.21$\times$ energy saving.

\end{abstract}
\section{Introduction}
\label{sec:intro}

The past decade has witnessed the tremendous development of convolutional neural network. It has achieved state-of-the-art performance on both low-level and high-level computer vision tasks. The outstanding capability comes from complicated network structure and massive computation. For example, ResNet-50 has 3.53G FLOPs and 25.56M parameters, neural image compression with residual learning~\cite{cheng2019deep} has more than 30G FLOPs and 111M parameters, which makes them difficult to operate on resource-strapped devices such as mobile phones and laptops.

For neural network computing, hardware devices need to obtain activation feature maps and filter parameters through bandwidth-limited on-off chip communication~\cite{samajdar2018scale,rhu2018compressing}. Such external memory access has impact on both energy consumption and inference latency, that is, it can become a bottleneck when data transfer latency is longer than computing~\cite{rhu2018compressing,zhe2021rate}. On the other hand, external memory access can cause 1000$\times$ more energy consumption than computation on microchips~\cite{horowitz20141}.
 
To deal with such obstacles, wight quantization~\cite{wang2019haq,girish2022lilnetx}, compression~\cite{girish2022lilnetx,oktay2019scalable} and coding~\cite{zhe2021rate} can be leveraged to compress the byte size needed for external memory access. However, in low-level tasks dealing with image nuances, the size of activation maps can be much larger than that of filter parameters. For example, in image compression network such as~\cite{balle2018variational}, the first convolutional layer of analysis transform has 3$\times$128$\times$5$\times$5 parameters, which equals to 38.4KB in 32bits float-point number; while a picture of Kodak dataset down-sampled by this layer has a dimension of 128$\times$256$\times$384, which equals to 50.3MB, nearly 1300 times larger than the weight parameters.

\begin{figure}[t]

    \centering
    \includegraphics[height=6cm]{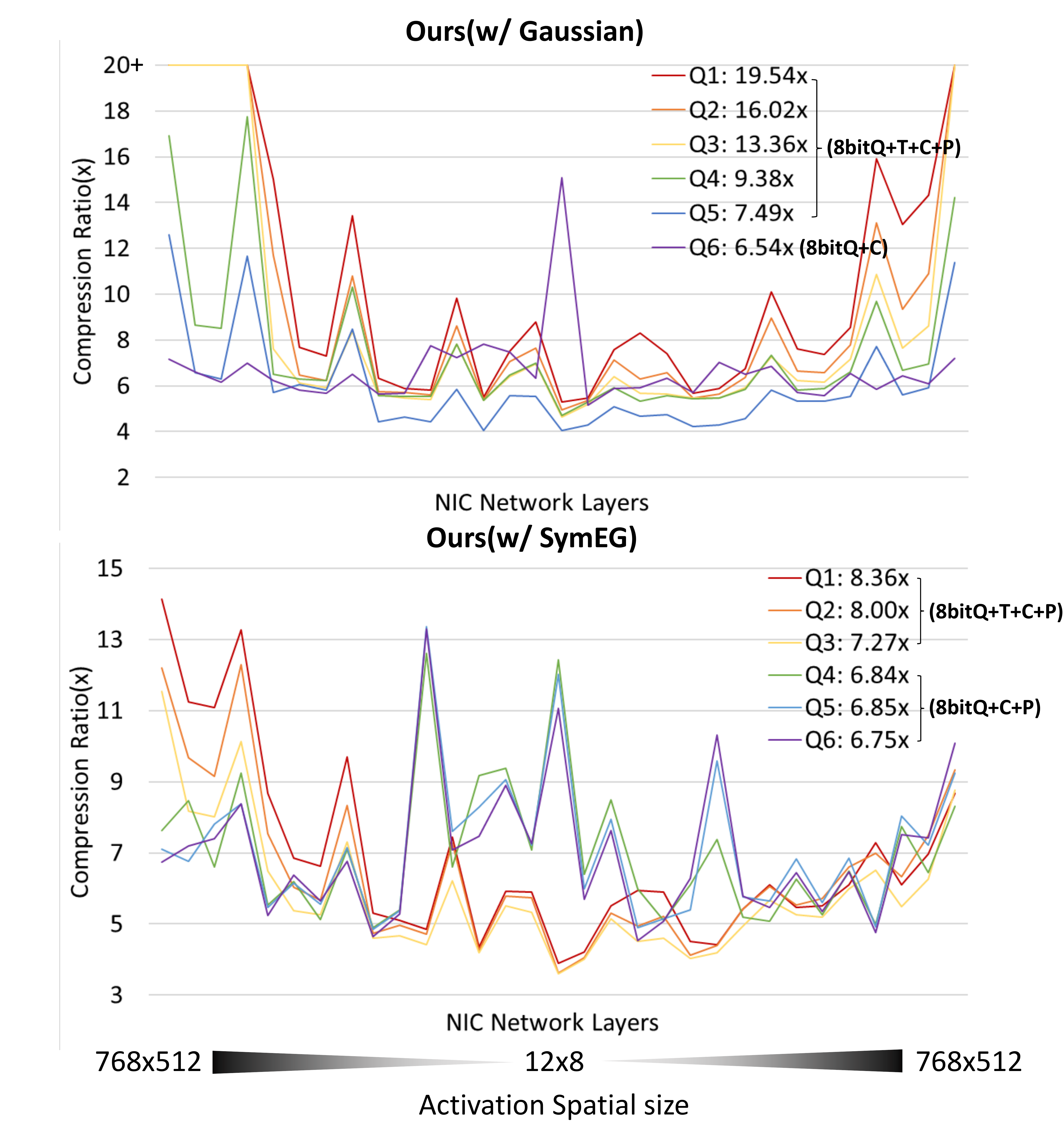}
    \caption{Layer-wise activation map compression ratio on Kodak. Compression is generally more obvious on larger activation maps, while turning off transform ingredient inverses this trend (Q4-Q6 of \textit{ours(w/ SymEG)} and Q6 of \textit{ours(w/ Guassian)}) for better RD performance.}
    \vspace{-0.3cm}
    
    \label{fig:1}

\end{figure}

\begin{figure*}[t]

    \centering
    \includegraphics[width=16.8cm, height=4.5cm]{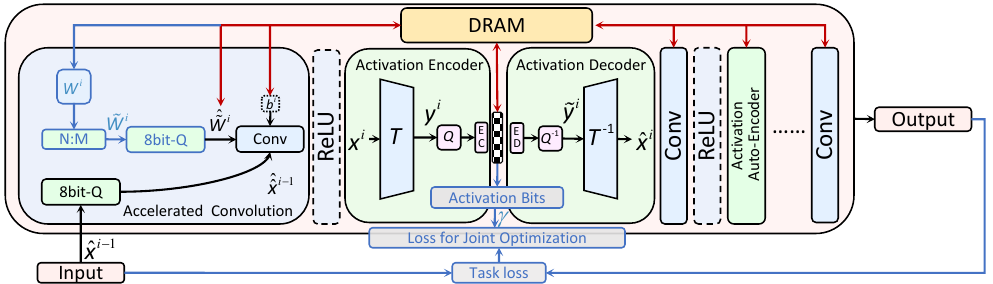}
    \caption{The proposed jointly-optimized bandwidth efficient convolutional neural network. \textit{EC} and \textit{ED} represent entropy encoding/decoding. Blue arrow or variables are only used in training. By compressing the activation in place, the access to DRAM is reduced.}
    \label{fig:2}

\end{figure*}

One possible solution is leveraging activation map compression for inference. Georgiadis~\cite{georgiadis2019accelerating} proposed a pipeline of sparsification, quantization and entropy coding (similar to the pipeline of pruning, quantization and coding~\cite{han2015deep} for weight compression) for convolutional network activation compression.) However, this pipeline  uses sparsity of activation maps as regulation for optimization, rather than the coded size of activation map directly and only reports its effectiveness on high-level image classification task.

Different from previous work, our method takes advantage of neural compression framework to make the first attempt to build a pipeline of transform, quantization and entropy coding for activation compression of low-level image compression task. Activation maps are first transformed to a latent space, then quantized to integer numbers. We propose a symmetric exponential Golomb (SymEG) coding, and a data-dependent Gaussian entropy model to calculate the real byte size of activation maps or their entropy proxies and using the total size of activation maps as a penalty term. Finally, we integrate \textit{int8} convolution~\cite{vanholder2016efficient} and $\textit{n:m}$ weight sparsity~\cite{zhou2021learning} to form the complete end-to-end jointly optimized inference-friendly pipeline with both efficient computation and efficient external memory access.

\section{Method}
\label{sec:Method}

The architecture of the proposed jointly-optimized bandwidth efficient convolutional neural network is shown in Fig.~\ref{fig:2}, which combines efficient computation and external memory access. For convolutional operations, weight parameters can be \textit{n:m} sparsified and quantized into 8-bit unsigned integer with input feature. For the activation map, it can be transformed, quantized and entropy coded to reduce byte size for DRAM (off-chip memory) access. During training, The actual (or estimated) byte size is weighted by a trade-off factor and added to the original task loss for joint training.

\subsection{Activation Auto-Encoder}
With generality, let's regard the activation map of the $i_{th}$ convolutional layer as its original output (or subsequent output after ReLU if applicable), and denote it as $\textbf{x}^{i}\in R^{C^i \times H^i \times W^i}$.  The activation map is transformed to the latent space $\textbf{y}^{i}\in R^{C^i \times H^i \times W^i}$  by transform $T$, while its dimension remains unchanged.
\begin{equation}\label{eq1}
\textbf{y}^{i} = T(\textbf{x}^{i})
\end{equation}
Note that transform $T$ is used for data statistics decorelation~\cite{young2021transform} or re-distribution for subsequent entropy coding. It is a general component that can be linear, non-linear or any other format. We examine 1$\times$1 convolution in our work, denoted as $T_{conv1}$.

Then, the latent activation map is quantized to unsigned integer numbers with uniform quantization:
\begin{equation}\label{eq2}
\hat{\textbf{y}}^{i} = Q(\textbf{y}^i) =\lceil \frac{\textbf{y}^{i}-\textbf{y}^{i,min}}{\textbf{y}^{i,max}-\textbf{y}^{i,min} }\cdot (2^{q}-1)\rfloor
\end{equation}
in which, $ \textbf{y}^{i,min}$ and  $ \textbf{y}^{i,max}$ represent the minimum and maximum value in $\textbf{y}^{i}$, $q$ is the bit depth, and $\lceil \cdot \rfloor$ means rounding the number to the  nearest integer. After entropy decoding, it is transformed back to the original data space using $Q^{-1}$ and  $T^{-1}$:
\begin{equation}\label{eq3}
\tilde{y}^{i} = Q^{-1}(\hat{\textbf{y}}^{i} )=\frac{\hat{\textbf{y}}^{i} \cdot (\textbf{y}^{i,max}-\textbf{y}^{i,min})}{2^{q}-1}+\textbf{y}^{i,min}
\end{equation}
\begin{equation}\label{eq4}
\hat{\textbf{x}}^{i} = T^{-1}(\tilde{\textbf{y}}^{i} )
\end{equation}

Note that $T^{-1}$ and $T$ are independent and contain trainable parameters of their own. $\hat{\textbf{x}}^{i}$ then goes into next convolutional layer as input to generate activation map $\textbf{x}^{i+1}$:

\begin{equation}\label{eq5}
\textbf{x}^{i+1} = \textbf{w}^{i+1}\ast\hat{\textbf{x}}^{i} +\textbf{b}^{i+1}
\end{equation}
or
\begin{equation}\label{eq6}
\textbf{x}^{i+1} = ReLU(\textbf{w}^{i+1}\ast\hat{\textbf{x}}^{i}+\textbf{b}^{i+1})
\end{equation}
in which $\ast$ denotes the convolution operation.

\subsection{Entropy Coding on Activation Map}

\subsubsection{Arithmetic Coding w/ Data-dependent Gaussian} 
 Following the practice in neural image compression~\cite{balle2018variational}, we propose our first entropy coding method, i.e., arithmetic coding with data-dependent Gaussian entropy model. Different from pixel-wise Gaussian parameter generated by  hyper codec which needs extra computation~\cite{balle2018variational}, we use the statistics of the data itself for estimation. The estimated parameters are generated in a channel-wise manner and broadcasted to every pixel for bit estimation.
For the $c_{th}$ channel in $\textbf{y}^{i}\in R^{C^i \times H^i \times W^i}$, the mean of its data is:
\begin{equation}\label{eq7}
\mathbf{\mu}_{c,:,:}^{i} = \frac{\sum_{h=1}^{H}\sum_{w=1}^{W}y_{c,h,w}^i}{H^i\cdot W^i}
\end{equation}

The standard deviation is:
\begin{equation}\label{eq8}
\mathbf{\sigma}_{c,:,:}^{i} = \sqrt{\frac{\sum_{h=1}^{H^i}\sum_{w=1}^{W^i}(y_{c,h,w}^i-\mathbf{\mu}_{c,:,:}^{i})^2 }{H^i\cdot W^i-1}}
\end{equation}

Pixel-wise Gaussian parameters are generatd by broadcasting the channel-wise parameters to every element in the channel:
\begin{equation}\label{eq9}
\mathbf{\mu}^{i} = [ \mathbf{\mu}_{0,:,:}^{i},\mathbf{\mu}_{1,:,:}^{i}....\mathbf{\mu}_{C-1,:,:}^{i} ]
\end{equation}
\begin{equation}\label{eq10}
\mathbf{\sigma}^{i} = [ \mathbf{\sigma}_{0,:,:}^{i},\mathbf{\sigma}_{1,:,:}^{i}....\mathbf{\sigma}_{C-1,:,:}^{i} ]
\end{equation}

Then we feed the calculated entropy parameters to Gaussian models to generate pixel-wise probability:
\begin{equation}\label{eq11}
q_{\hat{\textbf{y}}^{i}} =\mathcal{N}  (\mathbf{\mu}^{i},\mathbf{\sigma}^{i})
\end{equation}
The bit sequence of activation map can be given by arithmetic coding. 

\subsubsection{Symmetric Exponential Golomb} Although arithmetic coding has almost optimal coding efficiency and well-established hardware implementation, its latency renders it unsuitable for real-time processing.  On the contrary, Exponential-Golomb coding~\cite{TEUHOLA1978308} has been widely used in real-time video coding~\cite{wen1998reversible,nargundmath2013entropy}. The code of Golomb coding can be derived directly from the value of coding symbols and has less dependency on statistics of the data. 
To this end, we propose symmetric exponential Golomb coding. Among all integer numbers to be coded, we first choose a statistics-dependent reference number and calculate the differences between other numbers and the reference. Then, to deal with the signed differences, we double their absolute values for all differences and plus one for the  originally negative differences. And these unsigned differences are coded by the first stage exponential Golomb coding. We tested mean, mode and median as reference number and obtain 4.89$\times$, 5.95$\times$, 6.00$\times$ acitvation compression ratio of tested NIC model on Kodak dataset.  So we choose the median of data as the reference number in later experiments.

\subsection{Joint Optimization}
Let us denote the loss function of the original task as $\mathcal{L}_{ori}$, and the total byte size of activation map or its differential surrogate as $\mathcal{A}^i \in R^{b^{i }\times c^{i }\times h^{i }\times w^{i }}$. As the channel dimension $c$ is relatively stable, we averagely penalize the activation size on other dimensions, and the loss function for joint optimization can be written as:
\begin{equation}\label{eq12}
\mathcal{L} = \mathcal{L}_{ori}+\gamma \sum_{i=l}^{L}\frac{\mathcal{A}^i}{b^i \cdot h^i \cdot w^i}
\end{equation}where $L$ is the total number of activation maps, and $\gamma$ is the penalty factor for activation compression. For arithmetic coding, we use its entropy surrogate  estimated by data-dependent Gaussian model, which is the cross entropy of the estimated distribution of $\hat{\textbf{y}}^{i}$ and its actual distribution:

 \begin{equation}\label{eq13}
\mathcal{A}^{i}_{Gaussian} = \mathbb{E}_{\hat{\textbf{y}}^{i}\sim p_{y}}\{log_2q_{\hat{\textbf{y}}^{i}}[\hat{\textbf{y}}^{i}]\}
\end{equation}
During training, we use additional uniform noise to be the proxy of uniform quantization~\cite{balle2016end}. 
For symmetric exponential Golomb coding, we can obtain coding length directly and use it as penalization:

 \begin{equation}\label{eq14}
\mathcal{A}^{i}_{SymEG} =
\begin{cases}
2 \lfloor log_2[2(\hat{y}^i - \hat{y}_{ref}^i)+1] \rfloor +1 &  \hat{y}^i >\hat{y}_{ref}\\ 
\\
2 \lfloor log_2[2( \hat{y}_{ref}^i-\hat{y}^i )+2] \rfloor +1 & \text{otherwise}
\end{cases}
\end{equation}
The gradient of uniform quantization and floor function$\lfloor \cdot \rfloor$ are estimated by straight-through-estimator (STE)~\cite{bengio2013estimating}.

By now, optimization for end-to-end differentiable bandwidth efficient inference is formulated. In our experiments, we evaluate our proposed pipeline on low-level image compression with its $\mathcal{L}_{ori}=R+\lambda D$ defined by Rate-Distortion optimization on input images.

\section{Experiments}
\label{sec:Experiments}
The tested NIC network is based on \cite{balle2018variational} with GDN modules replaced by cascade bottleneck residual blocks. We adopt the training framework from CompressAI~\cite{begaint2020compressai} and set $\lambda$=\{0.0018, 0.0035, 0.0067, 0.0130, 0.0250, 0.0483\} ( Q1-Q6). The penalty $\gamma$ for activation compression in equation (\ref{eq12}) is $10^{-5}$. We use mean square error as the distortion metric $D$. We train each model for 100 epochs with batch size 16, using Adam optimizer with $\beta_1=0.9$, $\beta_2=0.999$ and initial learning rate of $10^{-4}$. After 64 epochs, the learning rate is decayed to $5\times 10^{-5}$. $n:m$ weight sparsity is set as $2:4$. The baseline and bandwidth efficient models are trained with the same setting.

As for our pipeline, we train two sets of models, one set uses Gaussian entropy model (equation.\ref{eq13}) for activation byte size penalization and the other uses symmetric exponential Golomb (equation.\ref{eq14}), denoted as "ours(w/ Gaussian)" and "ours(w/ SymEG)", respectively.

\begin{table*}[t]
  \centering
   \renewcommand\arraystretch{0.75}
       \begin{tabular}{ccccccccc}
    \toprule
    \textbf{Methods} & \textbf{Quality} & \textbf{Q1}    & \textbf{Q2 }   & \textbf{Q3}    & \textbf{Q4}    & \textbf{Q5}    &\textbf{ Q6 }   & \textbf{BD-Rate} \\
    \midrule
    Baseline & RDloss &   0.360    & 0.508      & 0.708      &  0.975     & 1.308      & 1.603      &0  \\
    \midrule
    \multirow{4}{*}{Ours(w/ Gaussian)} & RDloss &   0.368    & 0.525      &0.742       & 1.015      & 1.413      &     1.812  & \multirow{4}{*}{\textbf{0.071\%}} \\
          & Bandwidth$\downarrow$ &  19.54$\times$     & 16.02$\times$       & 13.36 $\times$     & 9.38$\times$     & 7.49$\times$     & 6.51$\times$      &  \\
           & Bandwidth$\downarrow$(w/ rans.) &  19.16$\times$     &15.91$\times$       &13.30$\times$       & 9.33$\times$      & 7.47$\times$      & 6.54$\times$      &  \\
          & Energy$\downarrow$ &  6.21$\times$   & 6.15$\times$ &   6.09$\times$    &     5.92$\times$  &  5.79$\times$   &  5.70$\times$      &  \\
    \midrule
    \multirow{3}{*}{Ours(w/ SymEG)} & RDloss &  0.365     &  0.516     & 0.729      &1.011       & 1.408      &   1.803    & \multirow{3}{*}{\textbf{-0.066\%}} \\
          & Bandwidth$\downarrow$ &  8.36$\times$   & 8.00$\times$     &   7.27$\times$    &     6.84$\times$  &  6.85$\times$   & 6.75$\times$      &  \\
          & Energy$\downarrow$ &  5.86$\times$   & 5.83$\times$ &   5.77$\times$    &     5.73$\times$  &  5.73$\times$   &  5.72$\times$      &  \\
          
    \bottomrule
    \end{tabular}%
    \caption{Results of evaluation on image compression. "$\downarrow$" denotes the ratio of saving for both items and the same for below.}
  \label{tab:tab2}%
\end{table*}%

\begin{table}[t]
  \centering
   \renewcommand\arraystretch{0.85}
       \begin{tabular}{l|cc}
    \toprule
    \multicolumn{1}{c|}{\textbf{NIC Q2}} & \textbf{RDloss} & \textbf{Bandwidth$\downarrow$} \\
    \midrule
    Baseline & 0.508 & - \\
    Q & 0.508 & 4.00$\times$ \\
    Q+E   & 0.508 & 6.49$\times$ \\
    Q+E+P & 0.508 & 10.23$\times$ \\
    Q+E+P+T & 0.515 & 15.96$\times$ \\
    Q+E+P+T+C & 0.525 & 16.02$\times$ \\
    Q+L1norm+SpaEG & 0.508 & 4.91$\times$ \\
   
    \bottomrule
    \end{tabular}%
    \caption{Effectiveness of each component of our pipeline. 
    Q: 8bit \textit{Q}uantization of $\mathbf{y}$. E: \textit{E}ntropy coding. P: Gaussian entropy \textit{P}enalty with $\gamma=10^{-5}$. T: conv1$\times$1 activation \textit{T}ransform. C:  \textit{C}omputation with \textit{n:m} weight sparsity and \textit{int8} convolution. \textit{L1norm+SpaEG} is our implementation of previous method~~\cite{georgiadis2019accelerating}.}
  \label{tab:3}%
\end{table}%

\begin{table}[th]
\scriptsize
  \centering
  \renewcommand\arraystretch{0.85}
          \begin{tabular}{c|ccc|cc}
    \toprule
    \multirow{2}{*}{\textbf{Settings}} & \multirow{2}{*}{\textbf{Coding}} & \multicolumn{1}{c}{\multirow{2}{*}{\textbf{Quant}}} & \multirow{2}{*}{\textbf{Penalty}} & \multicolumn{2}{c}{\textbf{Performance}} \\
          &       &       &       & Bandwidth$\downarrow$ & RDloss \\
    \midrule
    \multirow{3}{*}{Penalty} & \multirow{3}{*}{Guassian} & \multirow{3}{*}{8bit} & $10^{-4}$& 18.99$\times$ & 1.378 \\
          &       &       & $10^{-5}$ & 8.12$\times$ & 1.353 \\
          &       &       & $10^{-6}$ & 6.08$\times$ & 1.363 \\
    \midrule
    \multirow{5}{*}{Coding} & Guassian & \multirow{5}{*}{8bit} & \multirow{5}{*}{$10^{-5}$} & 8.12$\times$ & 1.353 \\
          & EG~\cite{TEUHOLA1978308}&       &       & 4.01$\times$ & 1.367 \\
          & SymEG &       &       & 5.96$\times$ & 1.360 \\
          & SpaEG~\cite{georgiadis2019accelerating} &       &       &    $4.05\times$   &  1.358 \\
          & Hyperprior~\cite{balle2018variational} &       &       & $9.33\times$ &1.353  \\
    \midrule
    \multirow{3}{*}{Bit depth} & \multirow{3}{*}{Guassian} & 6bit  & \multirow{3}{*}{$10^{-5}$} & 8.65$\times$   &1.574  \\
          &       & 8bit  &       & 8.12$\times$ & 1.353 \\
          &       & 10bit &       & 7.50$\times$ & 1.311 \\
    \bottomrule
    \end{tabular}%
    \caption{Results of ablation study on penalty, entropy coding method and activation bit depth. We set $k=4$ for EG.}
  \label{tab:5}%
\end{table}%
We record the sizes of activation map from each layer and calculate the total amount on the whole data set. Then, the bandwidth reduction is the total size of uncompressed 32bit float-point activation maps divided by the total amount of compressed ones with entropy coding overhead. The overheads include channel-wise mean and standard deviation for Gaussian entropy model, or channel-wise reference number for symmetric Golomb coding. These overheads are recorded as 32bit float-point numbers. Additionally, we follow~\cite{horowitz20141} to calculation the hardware energy consumption, and the energy reduction is the ratio of estimated consumption of baseline models to that of our pipelines.

The results are shown in Table.~\ref{tab:tab2}. We can see significant bandwidth reduction up to 19.54$\times$,  and the corresponding energy saving up to 6.21$\times$. When implementing actual entropy coding~\cite{rans2014} with our data-dependent Gaussian entropy estimation( w/ rans), the compression ratio is quite close to our entropy estimation, verifying the effect of entropy penalization. For model performances, we manage to control the BD-Rate~\cite{calculation2001} fluctuation below 0.1\%. Layer-wise compression results are provided in Fig.~\ref{fig:1}. 

To illustrate the effectiveness of each component and the interaction between them, we implement several combinations on Q2 model, as shown in Table.~\ref{tab:3}. We can see that transform and penalization are rather effective for bandwidth reduction.  For comparison  with existing pipeline (L1norm+SpaEG, ~\cite{georgiadis2019accelerating}), we find that nearly no bandwidth reduction on quantized NIC models, which indicates sparse regulation may not be suitable for NIC feature maps.


In addition, we did ablation study with Q5 model, no efficient convolution, $T_{conv1\times1}$ and the results are shown in Table.\ref{tab:5}. For penalty setting, it's not surprising that the larger the penalty $\gamma$ is, the higher the bandwidth reduction is.
For coding methods, our Gaussian and symmetric exponential Golomb outperform most of others on bandwidth reduction with nearly the smallest RD loss in case of NIC, except for mean-scale hyperprior~\cite{balle2018variational} with unacceptable complexity for activation compression. For activation bit depth, increasing bit depth to 10bit can trade less loss at the expense of less compression, while 6 bit activation causes serious performance degradation. Considering all the aspects studied above, settings for our NIC evaluations are shown in Figure.~\ref{fig:1}. These settings are to balance the dual-objective Rate-Distortion optimization with efficient computation and external memory access for different model qualities.

\section{Conclusion}
\label{sec:Conclusion}

In this paper, we proposed a transform-quantization-coding activation compression pipeline for efficient bandwidth inference that is differentiable and can be jointly optimized with original task and efficient computation methods. Further, we proposed data-dependent Gaussian entropy model and symmetric Golomb coding for activation penalization.
We are the first to implement bandwidth neural inference on low-level task like image compression and achieve up to 19$\times$ bandwidth reduction and 6.21$\times$ energy saving on test set.


\newpage

\section{Supplementary Material}
\label{sec:Supp}
In this supplementary material, we give more analysis and experimental details to support the main body of our paper. 1) More discussion on hardware friendliness of our pipeline.
2) More detail for proposed techniques.
3) For image compression implementation, we give detailed R-D performances of our models and details of  hyperprior activation coding. We also further explain why we choose settings in Figure.\ref{fig:1} of main body to build "ours" pipeline.

\subsection{More discussion on hardware friendliness}
As we illustrated in the introduction part, the main purpose of our method is to compress the intermediate feature maps (activations) of deep neural networks (DNNs) for efficient inference. To achieve bandwidth reduction from activation compression, activation should be compressed before transmitted to off-chip memory. Many previous research has illustrated the possibility of such configuration. For DNN accelerator, \cite{samajdar2018scale} presents the data flow of DNN calculation, where filter parameters and input feature maps are fetched from DRAM to on-chip SRAM before calculation and output feature maps are sent back to DRAM  through SRAM. \cite{park2020grlc} gives more detailed structure with a couple of data compressor and decompressor for data coding, they locate on chip and is between on-chip buffer and off-chip memory. Similar structure for CPU-GPU computing scenario is provided in \cite{rhu2018compressing}.

\begin{figure}[tph]

    \centering
    \includegraphics[width=8.5cm]{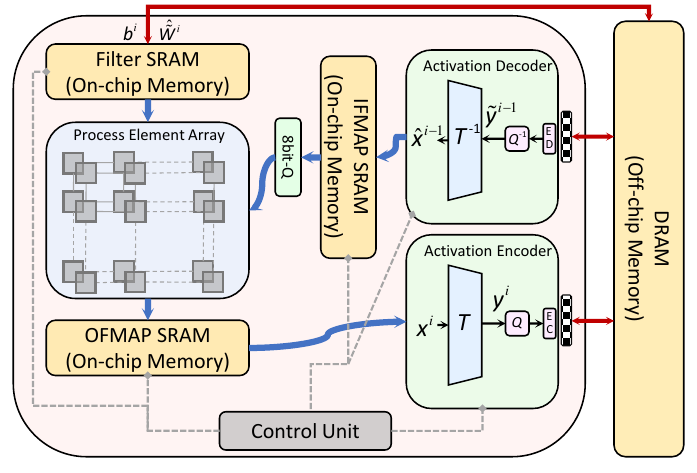}
    \caption{Detailed hardware data flow of our bandwidth efficient pipeline. The blue and red arrow represent the on-chip and external memory access respectively.  "IFMAP" and "OFMAP" denote input feature map and output feature map respectively.}
    \label{fig:s3}

\end{figure}

Following previous works, the hardware data flow of our pipeline considering the on-off chip communication are shown in Fig.\ref{fig:s3}.  The objective of our pipeline is to optimize external memory access that is denoted by red arrows, and the feature map are compressed/decompressed on-chip by activation en/decoder before being transmitted to/from DRAM.  Noted that in \cite{chmiel2020feature},  similar feature map transform coding are utilized, but it lacks direct penalization on the coded sized of activation. Besides, its PCA based transform matrix is pre-computed and re-calibrated time to time on pre-trained task models, which is orthogonal to our end-to-end jointly optimized pipeline.

\begin{figure}[t]

    \centering
    \includegraphics[width=7.5cm]{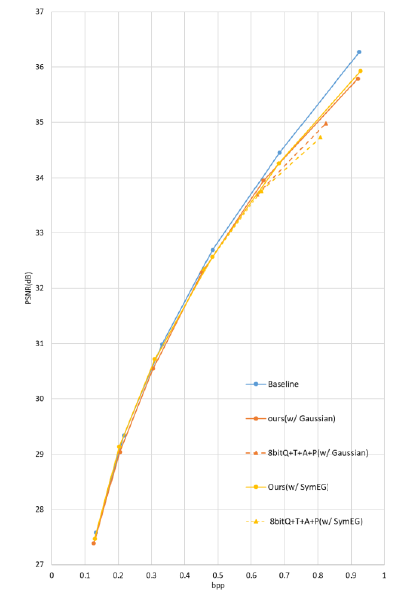}
    \caption{Comparison of R-D performances of ours setting and "8bitQ+T+A+P" setting .}
    \label{fig:s1}

\end{figure}

\subsection{More detail for proposed techniques}
\subsubsection{Detailed optimization with efficient convolution.}
Activation map compression mainly contributes to efficient external memory access. To further improve the practicability of our pipeline and improve computational efficiency, we add two optional ingredients, i.e., \textit{n:m} fine-grained weight sparsity~\cite{zhou2021learning} and \textit{int8} convolution. We follow ~\cite{zhou2021learning} to get \textit{n:m} sparsity weight and denote it as $\tilde{\textbf{w}}$, then the convolution operation in equation (\ref{eq5}) is modified as follows.

The scale of ${i+1}_{th}$ layer input $S_{input}^{i+1}$ is calculated by:
 \begin{equation}\label{eq15}
S_{input}^{i+1} = \frac{127}{max(abs(\hat{\textbf{x}}^i))}
\end{equation}
The scale of the ${i+1}_{th}$ layer weight $S_{weight}^{i+1}$ is calculated for per output channel by:
 \begin{equation}\label{eq16}
S_{weight,o}^{i+1} = \frac{127}{max(abs(\tilde{\textbf{w}}^{i+1}_{o,:,:,:}))}
\end{equation}
and with total $O$ output channels:
 \begin{equation}\label{eq17}
S_{weight}^{i+1} = [S_{weight,0}^{i+1},S_{weight,1}^{i+1}.....,S_{weight,O-1}^{i+1}]
\end{equation}

Then the input and weight of \textit{int8} convolution is:
 \begin{equation}\label{eq18}
\hat{\hat{\textbf{x}}}^i = \lceil S_{input}^{i+1} \cdot \hat{\textbf{x}}^i \rfloor_{-127\sim127}
\end{equation}
 \begin{equation}\label{eq19}
\hat{\tilde{\textbf{w}}}^{i+1} = \lceil S_{weight}^{i+1} \cdot \tilde{\textbf{w}}^{i+1} \rfloor_{-127\sim127}
\end{equation}
where $\lceil \cdot \rfloor_{-127\sim 127}$ means clamping to -127$\sim$127 after uniform quantization.
The output of \textit{int8} convolution is:
\begin{equation}\label{eq20}
\textbf{x}^{i+1} = \frac{\hat{\tilde{\textbf{w}}}^{i+1}\ast \hat{\hat{\textbf{x}}}^{i}}{S_{weight}^{i+1}\cdot S_{input}^{i+1}}+\textbf{b}^{i+1}
\end{equation}
The gradient of uniform quantization in equation (\ref{eq18},\ref{eq19}) is estimated by STE~\cite{bengio2013estimating}. During inference, the weights can be directly stored and accessed as $\tilde{\textbf{w}}^{i+1}$.

\subsubsection{Detail for Proposed Symmetric Exponential Golomb}

Naturally, the feature of exponential Golomb coding and Gaussian distribution can be combined by allocating code "1" to the most frequent number and trying to shorten the length of other numbers. To this end, we propose symmetric exponential Golomb coding (Algorithm.1). Among all integer numbers to be coded, we first choose a statistics-dependent reference number and calculate the differences between other numbers and the reference. Then, to deal with the signed differences, we double their absolute values for all differences and plus one for the  originally negative differences.And these unsigned differences are coded by the first stage exponential Golomb coding. We tested several reference numbers, as shown in Table.~\ref{tab1}. We can see that the mode and median of the to-be-coded data contribute to better compression ratio than the mean, while the median performs slightly better than the mode. So we choose the median of data as the reference number in later experiments.

\begin{table}[t]  
\centering

\renewcommand\arraystretch{0.9}
    \begin{tabular*}{\hsize}{@{}@{\extracolsep{\fill}}l@{}}  
        \toprule 
            \textbf{Algorithm 1} Symmetric Exponential Golomb Coding \\
        \hline  
            \quad \textbf{Input}: Unsigned integer $x$, reference number $x_{ref}$\\
            \quad \textbf{Output}: Bit stream $y$ and its length $l$\\
            \quad function \textbf{encode\_symmetric\_exp\_Golomb}($x$,$k$)\\
            \quad\{ \\
            \quad\quad $res$ = $x$-$x_{ref}$\\
            \quad\quad If \space$res<0$:\\
            \quad\quad\quad $z$ = 2 $\times$ abs($res$)\space+\space1\\
             \quad\quad Else :\\
             \quad\quad\quad $z$ = 2 $\times$ abs($res$)\\
             \quad\quad code = to\_binary($z\space+\space1$)\\
             \quad\quad $l_{code}$ = length(code)\\
             \quad\quad prefix = $\underbrace{0,0,....0}_{l_{code}-1}$\\
             \quad\quad $l$ =  $2 \times l_{code} - 1$\\
             \quad\quad $y$\space =\space concatenate(prefix, code)\\
             \quad \quad  Return $y$, $l$\\
             \quad\} \\
             \quad \textbf{Input}:  Bit stream $y$, reference number $x_{ref}$\\
            \quad \textbf{Output}: Unsigned integer $x$\\
            \quad function \textbf{decode\_symmetric\_exp\_Golomb}($y$)\\
            \quad\{ \\
            \quad\quad prefix, code = truncate\_before\_first\_1($y$)\\
            \quad\quad  $z$ = to\_integer(code)\space-\space1\\
            \quad\quad If \space$z\%2==0$:\\
            \quad\quad\quad $res$ = $z//2$\\
             \quad\quad Else :\\
             \quad\quad\quad $res$ = $ -1 \times (z-1)//2 $\\
             \quad \quad$x=res+x_{ref}$\\
             \quad \quad  Return $x$\\
             \quad\} \\
        \bottomrule  
    \end{tabular*}
    \label{alg1}
\end{table}
\begin{table}[t]  
\centering
\renewcommand\arraystretch{1.0}
    \begin{tabular}{cccc}
        \toprule 
        \textbf{Reference number} & Mean & Mode & Median \\
        \hline
       \textbf{Compression ratio} & 4.89$\times$ & 5.95$\times$& 6.00$\times$ \\
        \bottomrule  
    \end{tabular}
    \caption{Compression ratio of symmetric exponential Golomb with different channel-wise reference numbers on Kodak set~\cite{kodak}. Activation maps are obtained from activation 8-bit quantized baseline NIC model.}

    \label{tab1}
\end{table}

\subsection{More Results for Image Compression}
\subsubsection{Detailed R-D Performance of Neural Image Compression}
\begin{figure*}[tph]

    \centering
    \includegraphics[width=17cm]{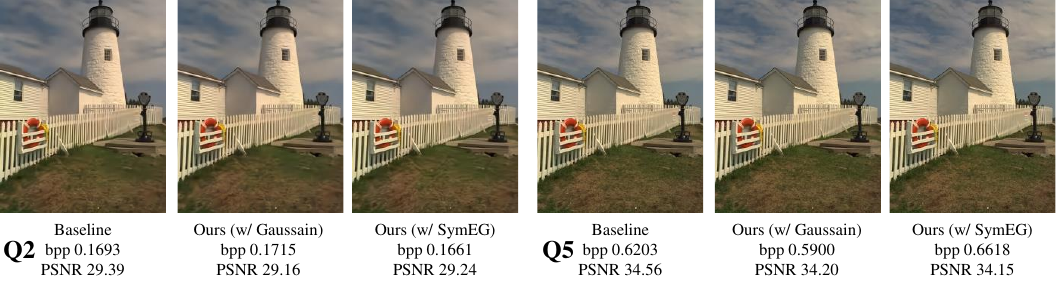}
    \caption{Subjective quality of Kodim.19 image from Kodak set. Two group of images are generated by Q2 and Q5 models respectively.}
    \label{fig:s4}

\end{figure*}

The detailed Rate-Distortion performance of our neural image compression(NIC) baseline models and two sets of bandwidth efficient NIC models are shown in Table.\ref{tab:s1}. We can see that we are able to cover the bit-per-pixel for an adequate range of 0.1 to 0.9 on Kodak test set\cite{kodak}. The example of subjective quality of reconstructed images are shown in Fig.\ref{fig:s4}. We can see that the reconstructions of our pipelines have little difference to that of baselines.

\begin{table*}[t]
  \centering

    \begin{tabular}{c|c|cccccc}
    \toprule
    \textbf{Model} & \textbf{Performance} & \textbf{Q1}    & \textbf{Q2}    & \textbf{Q3}    & \textbf{Q4}    & \textbf{Q5}    & \textbf{Q6} \\
    \midrule
    \multirow{2}{*}{baseline} & bpp   & 0.1334 & 0.2167 & 0.3315 & 0.4831 & 0.6848 & 0.9242 \\
          & PSNR  & 27.58 & 29.34 & 30.99 & 32.69 & 34.45 & 36.27 \\
    \midrule
    \multirow{2}{*}{Ours(w/ Gaussian)} & bpp   & 0.1266 & 0.2048 & 0.3044 & 0.4517 & 0.6364 & 0.9199 \\
          & PSNR  & 27.39 & 29.04 & 30.55 & 32.29 & 33.95 & 35.79 \\
    \midrule
    \multirow{2}{*}{Ours(w/ SymEG)} & bpp   & 0.1294 & 0.2027 & 0.3101 & 0.4825 & 0.6829 & 0.9284 \\
          & PSNR  & 27.47 & 29.14 & 30.72 & 32.57 & 34.26 & 35.93 \\
    \bottomrule
    \end{tabular}%
      \caption{Detailed R-D Performance of our NIC models}
  \label{tab:s1}%
\end{table*}%

\subsubsection{Explanation for our Components Setting}

\begin{figure}[t]

    \centering
    \includegraphics[width=7cm]{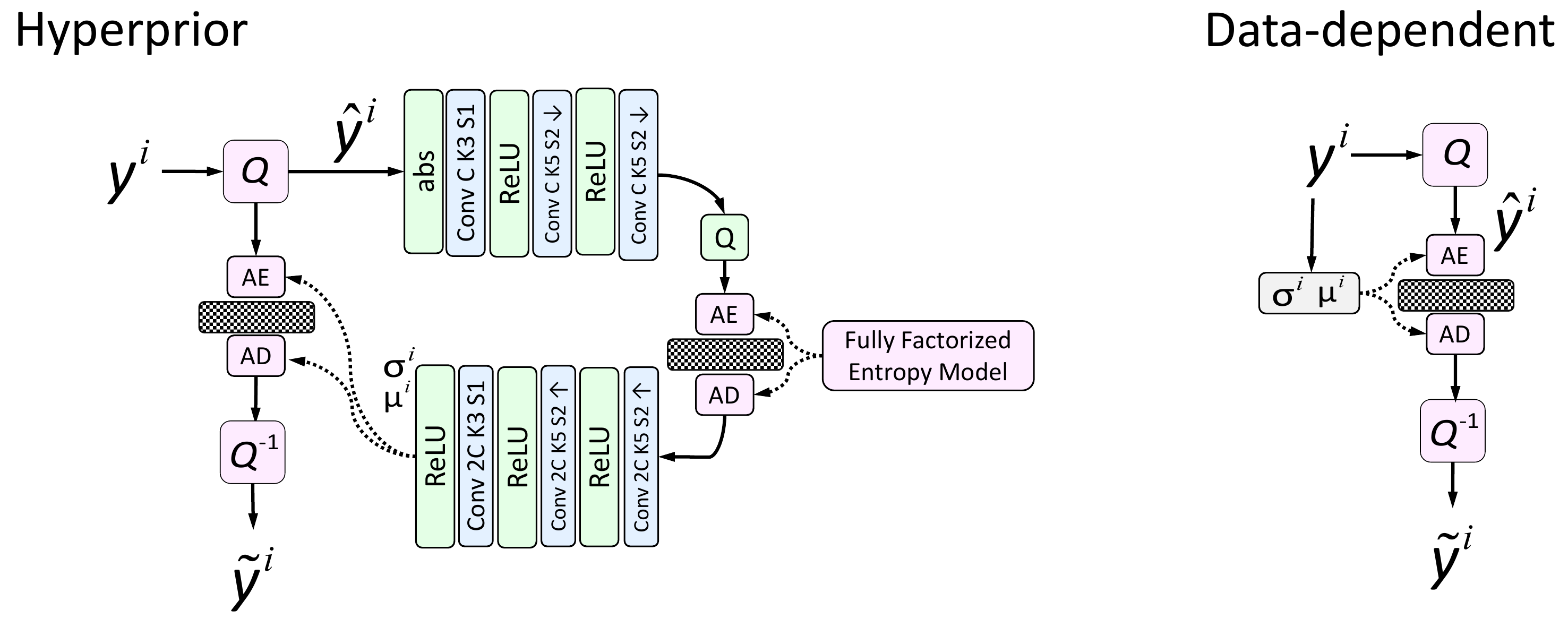}
    \caption{Comparison of Hyperprior  model(left) and our Data-dependent Gaussain model(right), $C$ denotes the channel number of $\textbf{y}^i$.}
    \label{fig:s2}

\end{figure}

\begin{table}[t]
  \centering
    \begin{tabular}{c|cc}
    \toprule
    \textbf{Model}  & \textbf{Ours(w/ Gaussian)} & \textbf{8bitQ+T+C+P} \\
    \midrule
    Q5     & 1.375, 7.49$\times$ & \textbf{1.413, 8.03$\times$} \\
    Q6   & 1.812, 6.51$\times$ & \textbf{1.983, 7.03$\times$} \\
    \bottomrule
    \end{tabular}%
     \caption{R-D loss and bandwidth reduction of ours(Gaussian) setting compared with "8bitQ+T+C+P" setting.}
  \label{tab:s2}%
\end{table}%

\begin{table}[t]

  \centering
    \begin{tabular}{c|cc}
    \toprule
    \textbf{Model}  & \textbf{Ours(w/ SymEG)} &  \textbf{8bitQ+T+C+P} \\
    \midrule
    Q4    & 0.989, 6.84$\times$ & \textbf{1.011, 6.61$\times$} \\
    Q5   & 1.342, 6.85$\times$ & \textbf{1.408, 5.84$\times$} \\
    Q6   & 1.803, 6.75$\times$ & \textbf{2.038, 5.45$\times$} \\
    \bottomrule
    \end{tabular}%
    \caption{R-D loss and bandwidth reduction of ours(SymEG) compared with "8bitQ+T+C+P" setting.}
  \label{tab:s3}%
\end{table}%
As shown in Table.4 of the main body, we use slightly different components setting of our schemes to ensure the Rate-Distortion performance of image compression task. In general, We use 8 bit activation quantization, $1\times1$ convolution transform, $2:4$ weight sparsity, $int8$ convolution and $\gamma=10^{-5}$ penalty for most of the models (denoted as "8bitQ+T+C+P"). Under this setting, we observe performance degradation for both schemes on higher quality models, 
as visualized in Fig.\ref{fig:s1}, where the dash lines with triangle marks represent the performance of "8bitQ+T+C+P" setting, while the solid line with dot marks represents the performance of "ours" setting.

This phenomena is rather intuitive, since higher quality NIC models have more delicate feature maps for less reconstruction distortion and need better preservation of image details. So their activations are harder to compress with simple transform like $1\times1$ convolution and their performances are more sensitive to bandwidth efficient inference. 

In Table.\ref{tab:s2} and Table.\ref{tab:s3}, we further verify our settings in Table.4 of the main body for several high quality models: 1) for "ours(w/ Gassian)", we use 10bit quantization instead of 8bit for Q5 model and we use identity transform without penalization for Q6 model. By doing so, We lost around 0.05 of bandwidth reduction for less RD loss. 2) for "ours(w/ SymEG)", we turn off $T_{covn1}$ for Q4-Q6 models, surprisingly achieving  more bandwidth reduction and less RD loss at the same time. This means $1\times1$ convolution transform is less effective for symmetric exponential Golomb coding, since it relies less on channel-wise distribution independence and can leverage the original Gaussian distribution of natural data statistics. To further stabilize training, for all "ours" models, we don't introduce weight sparsity and $int8$ convolution at the beginning, instead, during training, they are switched on after $10_{th}$ epoch and $30_{th}$ epoch respectively.

\subsubsection{Activation Coding with Hyperprior}
In section 4.1.3, we compare our entropy coding methods with other existing methods on NIC Q5 model. From the results, we can see that hyperprior\cite{balle2018variational} coding outperforms all methods in case of RD loss and bandwidth reduction. Hyperprior model is widely used in neural image compression to generate Gaussian parameters\cite{hu2021learning}, providing precise probability modeling for image entropy estimation and arithmetic coding. 
Specifically, we build mean-scale hyperprior\cite{minnen2018joint} to compare with our data-dependent Gaussian entropy model, as shown in Fig.\ref{fig:s2}. An additional hyper codec is applied on activation $\textbf{y}^i$, it is first down-sampled twice and tranformed to hyper latent by three cascade convolution layers, then inversely transformed to formulate Gaussian parameter $\mu^i$ and $\sigma^i$ with channel number doubled. The hyper latent is coded by arithmetic coding with fully factorized entropy model\cite{balle2018variational}. When calculating the bandwidth reduction, coded hyper latents are included in activation bits.

Despite the better compression brought by this pixel-wise Gaussian parameter generation, the computational cost of hyper codec is rather unsuitable and impractical for activation compression.  For a single activation map, it has six more convolution layers, a fully factorized entropy model containing four activated fully connected layers and two sets of arithmetic coding for both activation and hyper latent. Instead of generating Gaussian parameters in a pixel-wise manner with extra hyper codec, our method use channel-wise data statistics and only need mean and standard deviation calculation for each channel. Calculations are largely saved, while losing bandwidth reduction only from 9.33$\times$ to 8.12$\times$. The improvement brought by our data-dependent Gaussian is significant and more practical.


\bibliographystyle{IEEEbib}
\bibliography{refs}

\section{Copyright Statement}
\textit{© 2023 IEEE. Personal use of this material is permitted. Permission from IEEE must be obtained for all other uses, in any current or future media, including reprinting/republishing this material for advertising or promotional purposes, creating new collective works, for resale or redistribution to servers or lists, or reuse of any copyrighted component of this work in other works.}
\end{document}